\documentclass{article} 
\usepackage{iclr2023_conference,times}

\usepackage{amsmath,amsfonts,bm}

\def\eqref#1{equation~\ref{#1}}

\def\1{\bm{1}}

\DeclareMathAlphabet{\mathsfit}{\encodingdefault}{\sfdefault}{m}{sl}
\SetMathAlphabet{\mathsfit}{bold}{\encodingdefault}{\sfdefault}{bx}{n}

\usepackage{hyperref}
\usepackage{url}

\usepackage[utf8]{inputenc} 
\usepackage[T1]{fontenc}    
\usepackage{hyperref}       
\usepackage{url}            
\usepackage{booktabs}       
\usepackage{amsfonts}       
\usepackage{nicefrac}       
\usepackage{microtype}      
\usepackage{xcolor}         
\usepackage{amsmath} 
\usepackage{graphicx}
\usepackage{wrapfig}

\usepackage{amssymb}
\usepackage{pifont}
\newcommand{\xmark}{\ding{55}}
\newcommand{\cmark}{\ding{51}}

\definecolor{hollywoodcerise}{rgb}{0.96, 0.0, 0.63}
\definecolor{lasallegreen}{rgb}{0.03, 0.47, 0.19}
\definecolor{hanpurple}{rgb}{0.32, 0.09, 0.98}
\definecolor{green(pigment)}{rgb}{0.0, 0.65, 0.31}

\usepackage{hyperref}
\hypersetup{colorlinks,linkcolor={red},citecolor={hollywoodcerise},urlcolor={red}}

\title{Semi-Supervised Semantic Segmentation via Boosting Uncertainty on Unlabeled Data}

\newcommand{\conf}{{\text{\ttfamily{Conf}}} }

\author{Daoan Zhang, Yunhao Luo, Jianguo Zhang \\
Department of Computer Science\\
Southern Universty of Science and Technology\\
}

\iclrfinalcopy 
\begin{document}

\maketitle

\begin{abstract}
We bring a new perspective to semi-supervised semantic segmentation by providing an analysis on the labeled and unlabeled distributions in training datasets. We first figure out that the distribution gap between labeled and unlabeled datasets cannot be ignored, even though the two datasets are sampled from the same distribution. To address this issue, we theoretically analyze and experimentally prove that appropriately boosting uncertainty on unlabeled data can help minimize the distribution gap, which benefits the generalization of the model. We propose two strategies and design an uncertainty booster algorithm, specially for semi-supervised semantic segmentation. Extensive experiments are carried out based on these theories, and the results confirm the efficacy of the algorithm and strategies. Our plug-and-play uncertainty booster is tiny, efficient, and robust to hyperparameters but can significantly promote performance. Our approach achieves state-of-the-art performance in our experiments compared to the current semi-supervised semantic segmentation methods on the popular benchmarks: Cityscapes and PASCAL VOC 2012 with different train settings.
\end{abstract}

\section{Introduction}

Semantic segmentation \cite{zhang2023rethinking, wang2023prototype, wang2023bootstrap} has been a fundamental tool for various downstream applications. When deep learning methods are adopted in this area, the lack of fine-grained annotations is gradually prominent. Our paper focuses on semi-supervised semantic segmentation\citep{chen2021semi,luo2021ctbct,luo2021dtc,luo2021urpc}, which learns a model with a few labeled data and excess unlabeled data. Under these settings, how to appropriately utilize unlabeled data to improve generalization becomes critical \cite{wang2023feature, zhang2023aggregation, li2022transvlad, li2023cross}. 

Notice that, even if all labeled and unlabeled data are sampled from the same distribution, there is still a non-negligible distribution gap between the two clusters of data\cite{wang2023black}. This is the key question we deal with in this paper. Some recent approaches have attempted to tackle this question by designing consistency regularization\citep{chen2021semi,lee2021anti,luo2021dtc} or evaluating unlabeled \textit{o.o.d} (out of distribution) data via uncertainty approaches\citep{wang2022semi,kwon2022semi}, when traditional methods always reduce the output uncertainty to get an improvement. However, we want to argue that, due to the distribution gap, boosting uncertainty on the logits of unlabeled \textit{o.o.d} data can benefit the generalization of the model in semi-supervised semantic segmentation. 

We theoretically prove that elaborately designing an uncertainty booster for the model and applying it to unlabeled data can reduce the distribution gap, which can improve the generalization of the model. After that, we propose the requirements and strategies to design a suitable uncertainty booster for segmentation. The core principle is that we should consider the original distribution of the unlabeled images. Specifically, we demonstrate two strategies of selecting proper distribution and proper \textit{o.o.d} data. 

Based on the proposed strategies, we design an uncertainty booster for semi-supervised semantic segmentation to alleviate the distribution gap between the labeled and unlabeled datasets. Our newly designed module is benefited from the following advantages:

(1) \textbf{Plug-and-play} The uncertainty booster can be used in all semi-supervised segmentation methods that require retraining pseudo labels.

(2) \textbf{Green and Efficient} There are no trainable parameters and only a few fixed parameters in the booster, which means our module imposes nearly no impact on the training speed and only takes up very little memory and training sources.

(3) \textbf{Robustness} Due to our ablation study in Table \ref{t4}, our module is very robust to hyperparameters. Different hyperparameters show little influence on the promotion.

Experiments are carried out on the chosen baseline in \citep{chen2021semi}. Our strategy achieves state-of-the-art performance compared to current methods on the Cityscapes\citep{cordts2016cityscapes} and PASCAL VOC 2012 \citep{pascal-voc-2012} benchmarks under various data partition protocols.

\subsection{Related Work}

\paragraph{Semi-supervised learning} Semi-supervised learning has two typical paradigms: consistency regularization\citep{bachman2014learning,french2019semi,sajjadi2016regularization,xu2021dash} and self-training\citep{lee2013pseudo,zou2020pseudoseg}. The derived methods focus on data augmented self-training which utilizes strong augmentation such as CutMix\citep{yun2019cutmix}, CutOut\citep{devries2017improved}, ClassMix\citep{olsson2021classmix}. Recent approaches pay attention to how to better release the potential of unlabeled data\citep{mendel2020semi,ke2020guided,kwon2022semi,wang2022semi}, which, for example, aim to improve the quality of pseudo labels via distinguishing reliable and unreliable pseudo label\citep{wang2022semi}. However, these methods do not theoretically analyze the difference between the distributions of unlabeled and labeled datasets, which is the essence of making full use of unlabeled data. In contrast, our method gives a complete analysis of this question and designs an algorithm for semi-supervised semantic segmentation. 

\paragraph{Uncertainty in Deep learning} Uncertainties\cite{wang2023towards} can be divided into \textit{aleatoric uncertainty} and \textit{epistemic uncertainty}\citep{gal2016uncertainty}. The aleatoric uncertainty is also referred to \textit{data uncertainty}, which means some of the ground truth may be incorrect. The epistemic uncertainty, referred to as model uncertainty, represents the uncertainty of the model, including whether the model parameters best explain the observed data and whether the structure best fits the data. Some classical approaches that qualify uncertainty include bayesian epistemic uncertainty estimation via dropout\citep{gal2016dropout}, aleatoric uncertainty estimation via multi-network outputs\citep{kendall2017uncertainties}, epistemic uncertainty estimation via ensembling\citep{lakshminarayanan2017simple}. While for self-training in semi-supervised semantic segmentation, the pseudo labels of unlabeled data play the role of ground truth to finetune the model, which involves both aleatoric and epistemic uncertainty. We can connect both of the uncertainties by presenting high-quality pseudo labels. Thus, in this paper, we only consider the uncertainty of unlabeled data to analyze both of the uncertainties.

\section{Preliminaries}

\subsection{Boosting Uncertainty}
In this subsection, we will briefly introduce boosting uncertainty. Many common methods adopt minimizing uncertainty as an effective strategy to reduce overfitting, yielding better performances. A simple way to minimize uncertainty is adding $l_2$ regularization, forcing the model to produce a convincing result. While boosting uncertainty aims to let the model output a slightly fuzzy result and change the distribution, take a simple example, if the original output is a $[0.9, 0.05, 0.05]$ for a classification model, we may modify the model to yield $[0.85, 0.075, 0.075]$ instead of the original one via boosting uncertainty. In this case, the gap between the distribution of the model output after boosting uncertainty and the original distribution is getting wider hinges on the boosting strategy we choose.

\subsection{Settings}
Before we describe our findings, we shall clarify the symbols and settings used in this paper. To simplify the statement, we divide the ideal training dataset distribution \(D\) into two subsets, \(D_L\) and \(D_U\), respectively denoted as sampled distributions of the labeled and unlabeled datasets. We then denote the \(i.i.d.\) sampled elements of distribution \(D_L\) as \(L = \{(x_d, y_d)\}_{d=1}^{N} \sim D_L\) as the empirical distribution of \(D_L\), in which \(x\in \mathbb{R}^{k\times k}\) is from a \(k\times k\)-dimensional input space, and \(y\in\{1, 2, \cdots, K\}\) where \(K\) is the number of classes. The same applies to \(D_U\). The loss we use is \textit{0-1 loss}, defined as: \(\mathcal{L}(\cdot ) = \mathbb{1} \{h(x) \neq y\}\).

We first denote the vanilla model, which is trained only on the labeled dataset. Then we denote a posterior distribution $P$ on hypotheses that depends on the real distribution $L$ of the labeled dataset, parameterized by
$\mathbf{W}_l$ with bounded induced norm: $P=\prod_{i=1}^{d} \mathcal{N}(\mathbf{W}_l \bar{x_l}, \mathbf{I})$, whereas $\mathbf{W}_l$ is the weights and biases of vanilla model $h$. In this settings, $\bar{x_l}$ is the input representations from $L$. 

Secondly, we denote another posterior distribution $Q$ on hypotheses that depends on dataset $D_U$ \textbf{and} distribution $P$, parameterized by
$\mathbf{W}_{lu}$: $Q=\prod_{i=1}^{d} \mathcal{N}(\mathbf{W}_{lu} \bar{x}_{u}, \mathbf{I})$, whereas $\mathbf{W}_{lu}$ is the weights and biases of $h$ trained on $L$ and $U$, $\bar{x_u}$ is the input representations from $U$. 

We finally denote an uncertainty boosted posterior distribution on hypotheses that also depends on dataset $D_U$ \textbf{and} distribution $P$:  $Q_m=\prod_{i=1}^{d} \mathcal{N}((\mathbf{W}_{lu}+ \mathbf{b}) \bar{x}_{u}, \mathbf{I})$,  parameterized by
$\mathbf{W}_{lu}$, which is the weights and biases of $h^\prime$, whereas $\mathbf{b}$ indicates the distribution of uncertainty booster.  
By this, we define $R_{D_U}^E(h)$ as the \textit{Expected Risk} of $h$ applying on $D_U$ and $R_{D_U}^G(h)$ as the \textit{Empirical Risk} of $h$ applying on $D_U$.



\section{Theoretical Motivation For Boosting Uncertainty on Unlabeled Data}
This section aims to figure out whether boosting uncertainty on unlabeled data can help the model improve the generalization in semi-supervised semantic segmentation. Even though the labeled and unlabeled datasets are sampled from the same distribution, there is still a non-negligible distribution gap between these two sub-distributions, which is harmful to the model to yield pseudo labels for the unlabeled dataset. This section aims to explore and give an effective solution to this question.
\subsection{Boosting Uncertainty Helps Remitting Distribution Gap Between Labeled and Unlabeled Data Distributions}

We will begin our derivation by considering the vanilla semi-supervised semantic model and its variant of using an uncertainty booster.   
Given a trained model on labeled data, we will first explore the difference between the expected and empirical risks of the two models(vanilla and the variant of using uncertainty booster) on unlabeled data.

Our theory aims to find a model that can perform well on both labeled and unlabeled datasets, which can generate a better pseudo label for the unlabeled dataset for further training. This means the model can reveal the distance between labeled and unlabeled distributions and thus have a good generalization. On this basis, the optimization goal for minimizing the distribution gaps is defined as:
\begin{equation}
\label{1}
  F_2(h, h^\prime, D_L, D_U) = \underset{h^\prime}{\min}|R_{D_U}^E(h^\prime) - R_{D_L}^E(h)|
\end{equation}
where $h^\prime$ is the model that utilizes the uncertainty booster. As $R_{D_L}^E(h)$ depends on the model and labeled dataset selection, we can regard the $R_{D_L}^E(h)$ as a constant. Thus, we mainly focus on how the uncertainty booster influence $R_{D_U}^E(h^\prime)$.

\paragraph{Theorem 1} \citep{mcallester2003simplified} and \citep{germain2016new} provide an upper bound of the difference of expected risk $R_{D_U}^E(h^\prime)$ and empirical risk $R_{D_U}^G(h^\prime)$ with probability of at least $1-\delta$:


\begin{equation}
\label{2}
  R_{D_U}^E(h^\prime) - R_{D_U}^G(h^\prime) \leq \sqrt{\frac{\mathcal{\mathbf{KL}}[Q_m \| P]+\log \frac{2 \sqrt{N}}{\delta}}{2 N}}
\end{equation}
where $\mathbf{KL}$ is the K-L divergence, $N$ is the number of the samples in $U$.




For a semi-supervised model that utilizes the uncertainty booster, the K-L divergence in the upper bound can be calculated as:
\begin{equation}
\label{3}
\begin{aligned}
\mathbf{KL}[Q_m \| P] &=\sum_{i=1}^{d} \mathbf{KL}(\mathcal{N}((\mathbf{W}_{lu}+ \mathbf{b}) \bar{x}_{u}, \mathbf{I}) \| \mathcal{N}(\mathbf{W}_l \bar{x_l}, \mathbf{I})) \\
&=d\|\mathbf{W}_l \bar{x_l}-(\mathbf{W}_{lu}+ \mathbf{b}) \bar{x}_{u}\|_{2}^{2}
\end{aligned}
\end{equation}

Therefore, from Eq. \ref{2} and Eq. \ref{3}, we have that:
\begin{equation}
\label{4}
  R_{D_U}^E(h^\prime) \leq R_{D_U}^G(h^\prime) + \sqrt{\frac{d\|\mathbf{W}_l \bar{x_l}-(\mathbf{W}_{lu}+\mathbf{b}) \bar{x}_{u}\|_{2}^{2}}{2N}} + 
  \sqrt{\frac{\log {\frac{2 \sqrt{N}}{\delta}}}{2 N}}
\end{equation}

Moreover, for the vanilla model $h$, the form of Eq. \ref{4} is written as:

\begin{equation}
\label{5}
   R_{D_U}^E(h) \leq R_{D_U}^G(h) + \sqrt{\frac{d\|\mathbf{W}_l \bar{x_l}-\mathbf{W}_{lu} \bar{x}_{u}\|_{2}^{2}}{2N}} + 
   \sqrt{\frac{\log {\frac{2 \sqrt{N}}{\delta}}}{2 N}}
\end{equation}


The detailed proof is presented in the appendix.



In Eq. \ref{1}, after training on labeled dataset $L$, the model can learn a sub-distribution on the distribution of the labeled dataset. As unlabeled data has no ground-truth labels, the \textit{Expected Risk}: $R_{D_U}^E(h)$ for predictor $h$ on unlabeled dataset distribution always surpass than that of $R_{D_L}^E(h)$ on labeled dataset. We will first discuss this most common situation.

\paragraph{Situation 1: $R_{D_U}^E(h^\prime)$ is larger than $R_{D_L}^E(h)$}

In this situation, Eq. \ref{1} turns out to minimize the upper bound of $R_{D_U}^E(h^\prime)$. This means we should minimize the RHS of Eq. \ref{4} compared to the RHS of Eq. \ref{5}. As the input data is the same and the influence of uncertainty booster is tiny in an iteration, we can suppose $R_{D_U}^G(h)$ and $R_{D_U}^G(h^\prime)$ remain the same. So the comparison of the scales of the upper bounds in the RHS of Eq. \ref{4} and Eq. \ref{5} turns out to focus on the scales of $\|\mathbf{W}_l \bar{x_l}-(\mathbf{W}_{lu}+\mathbf{b}) \bar{x}_{u}\|_{2}^{2}$

We can see that the aim of minimizing $\|\mathbf{W}_l \bar{x_l}-(\mathbf{W}_{lu}+\mathbf{b}) \bar{x}_{u}\|_{2}^{2}$ is to keep $(\mathbf{W}_{lu}+ \mathbf{b}) \bar{x}_{u}$ closer to $\mathbf{W}_l \bar{x_l}$ from labeled dataset. While for a uniform input $\bar{x}_u$ sampled from $U$, as the model $h$ has already trained on labeled dataset, the weights $\mathbf{W}_{lu}$ have a strong affinity for the distribution of labeled distribution $L$, thus when input \textit{o.o.d} data in unlabeled data, $\mathbf{W}_{lu} \bar{x}_u$ will be dragged far away from original labeled dataset distribution, which incurs a much higher upper bound of \textit{Expected Risk}, so we may push back $(\mathbf{W}_{lu}+ \mathbf{b}) \bar{x}_{u}$ to $\mathbf{W}_l \bar{x_l}$ via the appending of $\mathbf{b}$ which is an uncertainty booster that can slightly alter the distribution. Thus, if $\mathbf{b}$ is carefully designed and applied on possible unlabeled \textit{o.o.d} data, the $R_{D_U}^E(h^\prime)$ can have a lower upper bound than the original $R_{D_U}^E(h)$. We then further analyze how to design the proposed booster. 

\paragraph{Situation 2: $R_{D_U}^E(h)$ is smaller than $R_{D_L}^E(h)$} In this rare case, Eq. \ref{1} turns out to be:
\begin{equation}
\label{6}
  F_1(h, h^\prime, D_L, D_U) = \underset{h}{\min}(R_{D_L}^E(h) - R_{D_U}^E(h^\prime))
\end{equation}
Thus, we focus on maximizing the upper bound of $R_{D_U}^E(h)$, so in Eq. \ref{4}, we just simply introduce some random noise to increase the difference between $\mathbf{W}_l \bar{x_l}$ and $(\mathbf{W}_{lu}+\mathbf{b}) \bar{x}_{u}$ via $\mathbf{b}$ and hence, we can minimize function $F$.

In all, if better selected and designed, boosting uncertainty on unlabeled \textit{o.o.d} data may reduce the difference between the \textit{Expected Risks} of labeled and unlabeled distributions, indicating that this strategy helps minimize the potential distribution gap between labeled distribution and unlabeled distribution.

\subsection{A Theory of Designing Uncertainty Booster for Segmentation}
In the last subsection, we figure out that boosting uncertainty in segmentation may help reduce the distribution gap for labeled and unlabeled distributions. At the same time, there is still a disturbing risk when rethinking Eq. \ref{1}. Since boosting uncertainty may reduce the upper bound of $R_{D_U}^E(h^\prime)$, there are questions about how much we reduce the upper bound suitable for the model. If a lousy booster is chosen, was there a catastrophic influence on the distribution? We will further discuss these several problems.

\subsubsection{Conditions of Compliance when Boosting Uncertainty}
In this subsection, to better understand the dense segmentation task, we will focus on the pixel distribution in the image. We will begin with how to generate an excellent pseudo label for a single image.

\paragraph{Theorem 2} \citep{mcallester2003simplified} observed that, let $\mathcal{H}$ be a hypothesis space, $h \in \mathcal{H}$, $D_L$ be the labeled dataset distribution, $I_U$ be the distribution of a single unlabeled image. $\mathrm{R}_{X}(h)$ is the \textit{Expected Risk} on $I_U$, $\mathrm{R}_{D}(h)$ is the \textit{Expected Risk} on $D_L$. We have:

\begin{equation}
\label{7}
\forall h \in \mathcal{H}, \mathrm{R}_{X}(h) \leq\mathrm{R}_{D}(h)+\frac{1}{2} d_{\mathcal{H} \Delta \mathcal{H}}\left(D_L, I_U\right)+\mu_{h^{*}} 
\end{equation}

whereas the $d_{\mathcal{H} \Delta \mathcal{H}}$ is denoted as empirical discrepancy distance:

\begin{equation}
\label{8}
\begin{aligned}
&d_{\mathcal{H} \Delta \mathcal{H}}\left(D_L, I_U\right)= 2 \sup _{\left(h_1, h_2\right) \in \mathcal{H}^{2}}\left|\underset{\mathbf{x} \sim D_L}{\mathbb{E}} \mathcal{Pr} \left[h_1(\mathbf{x}) \neq h_2(\mathbf{x})\right]- \underset{\mathbf{x} \sim I_U}{\mathbb{E}} \mathcal{Pr}\left[h_1(\mathbf{x}) \neq h_2(\mathbf{x})\right]\right|
\end{aligned}
\end{equation}

and 
\begin{equation}
\label{9}
\mu_{h^{*}}=\mathrm{R}_{\mathcal{S}}\left(h^{*}\right)+\mathrm{R}_{\mathcal{T}}\left(h^{*}\right), h^{*}=\operatorname{argmin}_{h \in \mathcal{H}}\left(\mathrm{R}_{\mathcal{S}}(h)+\mathrm{R}_{\mathcal{T}}(h)\right)
\end{equation}
and $x$ is the pixels in the images.



As the unlabeled dataset has no ground-truth labels, the $\mu_{h^{*}}$ is inaccessible, and both the labeled dataset and the unlabeled image are sampled from the same distribution $D$. The $\mu_{h^{*}}$ is assumed to be low and trivial. As the target is to minimize the discrepancy between $\mathrm{R}_{X}(h)$ and $\mathrm{R}_{D}(h)$ in Eq. \ref{7}. The aim turns out to minimize $d_{\mathcal{H} \Delta \mathcal{H}}\left(D_L, I_U\right)$. Based on \citep{mansour2009domain}, we can modify Eq. \ref{8} into:

\begin{equation}
\label{10}
\begin{aligned}
d_{\mathcal{H} \Delta \mathcal{H}}\left(D_L, I_U\right) 
&=2 \sup _{\left(h_1, h_2\right) \in {\mathcal{H}}^{2}}\left|\underset{\mathbf{x} \sim {D_L}}{\mathbb{E}} \mathcal{L}_r\left(h_1(\mathbf{x}), h_2(\mathbf{x})\right)-\underset{\mathbf{x} \sim {I_U}}{\mathbb{E}} \mathcal{L}_r\left(h_1(\mathbf{x}), h_2(\mathbf{x})\right)\right| 
\end{aligned}
\end{equation}

Where $\mathcal{L_r}$ can be a general real-valued loss. 
If $L_r$ is a $L_2$ loss:

\begin{equation}
\label{11}
\begin{aligned}
d_{\mathcal{H} \Delta \mathcal{H}}\left(D_L, I_U\right) 
&=2 \sup _{\left(h_1, h_2\right) \in {\mathcal{H}}^{2}}\left|\underset{\mathbf{x} \sim {D_L}}{\mathbf{E}} \left[\left(h_1(x)-h_2(x)\right)^{2}\right]-\underset{\mathbf{x} \sim {I_U}}{\mathbf{E}} \left[\left(h_1(x)-h_2(x)\right)^{2}\right]\right| 
\\
&=2 \sup _{\left(h_1, h_2\right) \in {\mathcal{H}}^{2}} \left|\sum_{\mathbf{x} \in ({D_L} \cup {I_U})}\left[{D_L}(\mathbf{x})-{I_U}(\mathbf{x})\right] \left[(h_1(x)-h_2(x))^{2}\right] \right|
\end{aligned}
\end{equation}

In which we can see, $\left[(h_1(x)-h_2(x))^{2}\right]$ depends on $({D_L} \cup {I_U})$ which is sampled from the support of $D$ which we could hardly control and is greater than or equal to 0, so minimizing difference of $\mathrm{R}_{X}(h)$ and $\mathrm{R}_{D}(h)$ depends on minimizing $\left[{D_L}(\mathbf{x})-{I_U}(\mathbf{x})\right]$. This means for each pixel in unlabeled images, the difference between the pixel distribution of the output of the unlabeled images and the prior distribution of the unlabeled images should be as close as possible.

 

\subsubsection{Two Strategies for Designing an Uncertainty Booster for Segmentation}\label{subsec:2stra}
\paragraph{Strategy 1: The Criteria for Distribution Imitation} Based on Theorem 2, the uncertainty boosted output should have a similar distribution to the prior distribution of the image. From an intuitive perspective, the pixel distributions are various in different images, so we shall focus on image-wise distribution for each pixel when boosting uncertainty for a segmentation model. In a nutshell, \textit{the distribution of uncertainty booster that we apply to each image in the segmentation model shall be subject to the distribution of the image itself.}

\paragraph{Strategy 2: The Criteria for Data Selection} In addition to considering the selection criteria of the distribution, we shall also consider the scale that we boost uncertainty. We should focus on the unlabeled \textit{o.o.d} data relative to the prior output distribution. We first propose that if a particular model $h$ is trained on a known sampled distribution, we test it on a sampled data point. If the trained model yields a higher uncertainty on the sampled data point, the more likely this data point is out of distribution from the known distribution; thus, we need to give this data point a more significant disturbance. Based on Theorem 2, we raise the second template for designing the uncertainty booster: \textit{The scale of boosting uncertainty for data depends on the scale of the uncertainty that the model generates on the data.} This means that the model should pay more attention to the uncertain data points. The more the model is unconfident, the more we boost the uncertainty.

\section{Uncertainty Booster Module (UBM)}

In this section, based on the two strategies proposed in section \ref{subsec:2stra}, we design a plug-and-play uncertainty booster module (UBM) specialized for semi-supervised semantic segmentation. In the meantime, our proposed module requires negligible extra memory or computation while achieving noticeable performance gain for segmentation models.

\subsection{Regional Uncertainty Voter (RUV)}

We aim to find a proper distribution to boost pixel-level uncertainty according to Strategy 1. As mentioned above, commonly used Gaussian or Uniform Distributions applied to the whole dataset may fail in semantic segmentation because the distribution of a specific image varies from one another. The nature of semantic consistency renders a pixel closely related to its adjacent pixels in an image. Thus, imposing non-regional perturbation on pixels prohibits the model from learning the actual distribution. As such, we design the uncertainty booster on an image-wise case-by-case basis. 

We consider taking regional information into account and propose the \textit{Regional Uncertainty Voter} (RUV) to produce a customized artificial distribution.
Given a one-hot pseudo label $p^{oh} \in \mathbb{R}^{h \times w \times K}$ of an image $x^{h \times w}$, where $K$ is the number of classes. 


We count the number of pixels belonging to each class in the $h_v \times w_v$ vicinity $V$ of every individual pixel $x_{i}$, $i \in [0, hw-1]$, by which 
pixels can 
The module can perceive and aggregate unique regional information in the image, which is done by a specifically defined kernel.

Then we divide the counting result map by the cardinality of $V$ to yield a probability map $C \in \mathbb{R}^{h \times w \times K} $. 

Compared with the universal Gaussian or Uniform Distribution, our region-aware distribution is more natural and sensible to impose. We formulate the calculation of $C$ in Eq. \ref{eq:count}:
\begin{equation}
    C_j = \frac{\sum_{k=0}^{K-1}  \text{weight}(j, k) \star p^{oh}_{k} } { |V|} , \quad \text{where} \; \:
    \text{weight}(j,k) = \left\{
                                \begin{aligned}
                                    & \mathbf{1}^{h_v \times w_v} \quad \text{if} \: j = k  \\
                                    & \mathbf{0}^{h_v \times w_v} \quad \text{else} \\
                                \end{aligned}
                                \right. 
    \text{and} \ C_j \in C
    \label{eq:count}
\end{equation}
where $\star$ is the valid 2D cross-correlation operator, $C_j \in \mathbb{R}^{h \times w}$ is the probability map of class $j$, weight$(j,k)$ maps the $k$ th layer ($k$ th class) of $p^{oh}$ to $C_j$. Note that weight$(j,k)$ is a constant kernel for gathering neighbor predictions. The voter in Eq. \ref{eq:count} can be efficiently computed by {\ttfamily{Conv2d}} with our pre-designed kernel weight and is free from back-propagation, rendering neglectable computation cost. 


\subsection{Uncertainty Adaptive Strategy (UAS)}
Our uncertainty booster is required to be careful and smart. Intuitively, boosting uncertainty wildly would negatively impact performance because correct and certain distribution that already learned is likely to be deviated by the booster. Therefore, the selection criterion is crucial and should be tailored for each pixel of every image at every single state of training. To address this issue, we propose calculating the confidence value, {\ttfamily{Conf}}, based on the entropy of each pixel, which decides how strong the booster should be for the corresponding pixel. Conventionally, we regard pixels with great \conf value as well-classified ones, where extra uncertainty is unnecessary. While those with small \conf values are expected to be unconfident \textit{o.o.d} pixels, thus, a strong booster is needed. To sum up, assume $pred \in \mathbb{R}^{h \times w \times K}$ to be the prediction probabilistic map of the model, we define \conf of pixel $x_i \in x$ in Eq. \ref{eq:conf} and the normalized adaptive weight $W_{x_{i}} \in W_x$ of pixel $x_i$ for the pseudo label $p^{oh}$ in Eq. \ref{eq:entr_weight}:
\begin{equation}
    \conf (x_{i}) = \sum_{k=0}^{K-1}  pred_{i,k} \ \log  (pred_{i,k})
    \label{eq:conf}
\end{equation}

\begin{equation}
    W_{x_{i}} = \frac{\conf(x_{i}) - \min \conf(x) }{ \max \conf(x) - \min \conf(x) },  \ W_{x_{i}} \in W_x
    \label{eq:entr_weight}
\end{equation}

In all, given a vanilla pseudo label $p^{oh}$, based on the RUV and UAS, we can define the uncertainty boosted pseudo label $ \hat{p}_{i}$ in Eq. \ref{eq:sm_label}:
\begin{equation}
    \hat{p} =  \text{UBM}(p^{oh}) =  W_{x} * p^{oh} + (1 - W_{x}) * C,  \ p^{oh} = Onehot(Pred)
    \label{eq:sm_label}
\end{equation}

\begin{wrapfigure}{r}{0.65\textwidth}
 \vspace{-20pt}
 	\centering
 	\includegraphics[width=0.65\textwidth]{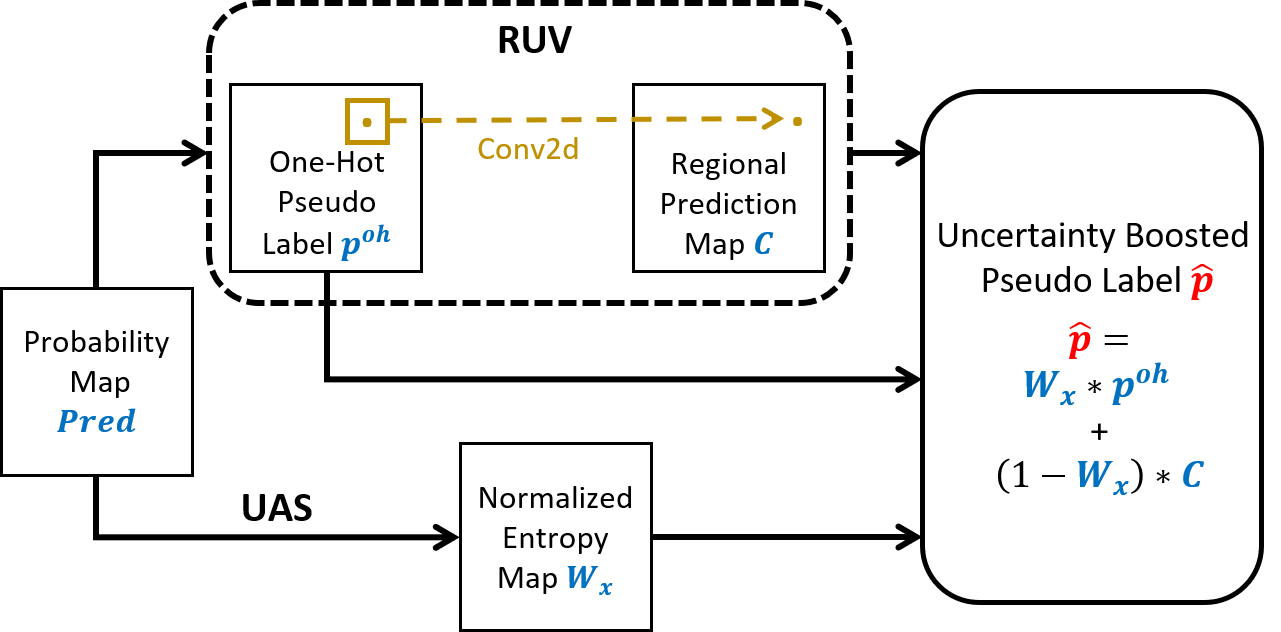} 
 	\caption{The pipline of UBM.}
 	\label{fig:main}
 	\vspace{-5pt}
 \end{wrapfigure}

The proposed pipeline is shown in Fig. \ref{fig:main}; the proposed UBM module is simple, low-parameters, and efficient. After the UBM module finishes processing the vanilla probability map $Pred$, we use the output pseudo labels $\hat{p}$ for further training.


\subsection{Overall Strategy}
We follow CPS \citep{chen2021semi} as the baseline, which consists of two independent models, namely $f(x; \theta_1)$ and $f(x; \theta_2)$. The two models have the same network structure and loss definition but different initialization.
For labeled data $L$, both $f(x; \theta_1)$ and $f(x; \theta_2)$ are trained by CrossEntropy (CE) loss with ground truth; For unlabeled data $U$, the models generate pseudo labels $p^\text{U}_1$ and $p^\text{U}_2$ for each other as the ground truth of CE loss. We only apply the proposed module UBM on the two pseudo labels $p^\text{U}_1$ and $p^\text{U}_2$. Finally, the overall loss function for $f(x; \theta_1)$ is defined in Eq. \ref{eq:loss}, vice versa for $f(x; \theta_2)$.
\begin{equation}
    \mathcal{L}_1  =  \frac{1}{|L|} \sum_{x \in L} \text{CE} (p, y)  +   \lambda \frac{1}{|U|}  \sum_{x \in U} \text{CE} (p^\text{U}_1, \text{UBM}(p^\text{U}_2))
    \label{eq:loss}
\end{equation}
where $p$ is the probabilistic output of labeled data, $y$ is the ground truth of labeled data, $p^\text{U}_1$ is the probabilistic output of the unlabeled data of this model, $p^\text{U}_2$ is the one-hot output of the same unlabeled data of the other model, UBM is the proposed uncertainty booster module. $\lambda$ is the trade-off weight between the two CE losses. 

\section{Experiments}
\paragraph{Datasets \& Evaluation} PASCAL VOC 2012 dataset is a standard object-centric semantic segmentation dataset, which contains 20 foreground classes and one background class. We follow previous works and adopt the augmented set \citep{hariharan2011semantic} with 10,582 training images and 1449 validation images. Cityscapes~\citep{cordts2016cityscapes} is a dataset for urban scene understanding, consisting of 2,975 training images with fine-annotated labels and 500 validation images. For both datasets, training images are split under label ratios of 1/16, 1/8, 1/4, and 1/2, respectively. We directly adopt all the split partitions provided by CPS~\citep{chen2021semi}.

 We use the mean Intersection-over-Union (mIoU) metric to evaluate the segmentation performance. We report results on PASCAL VOC 2012 val set and Cityscapes val set for all label ratios. Following \citep{wang2022semi}, for PASCAL VOC 2012, we center crop images to a fixed resolution; as for Cityscapes, we use sliding-window evaluation.

\paragraph{Implementation}
 We use ResNet-101\citep{he2016deep} pretrained on ImageNet\citep{deng2009imagenet} as our backbone and DeepLabv3+\citep{chen2018encoder} as the segmentation head. Following CPS~\citep{chen2021semi}, we add a deep stem block to our backbone, remove the last down-sampling operations, and employ dilated convolutions in the subsequent convolution layers. In addition, we use mini-batch SGD with the momentum of 0.9 and a weight decay of 0.0001 to train our model. As in CPS~\citep{chen2021semi}, we train PASCAL VOC 2012 for 60 epochs with $\lambda = 1.5$ and Cityscapes with OHEM loss for 240 epochs with $\lambda = 6$, where the learning rates are 0.0025 and 0.005, respectively. For simplicity, the size of the vicinity $V$ in UBM is set to $5 \times 5$ in all experiments.
 
\subsection{Comparison with Existing Alternatives}
We compare our method with the current state-of-the-art methods, including DCC\citep{lai2021semi}, ST++\citep{yang2022st++}, $\text{U}^2\text{PL}$\citep{wang2022semi} etc. Experiments are carried out via the same network architecture. Notice that we don't use CutMix\citep{yun2019cutmix}, which is a powerful augmentation, and we still achieve state-of-the-art performance.

\paragraph{Results on PASCAL VOC 2012} Table \ref{t1} compares our proposed method with state-of-the-art methods on PASCAL VOC 2012 dataset. Compared to the baseline, our uncertainty booster steadily promotes the performance, achieving impressive improvements of \textcolor{red}{\textbf{+1.41, +1.42, +1.75}} and \textcolor{red}{\textbf{+2.00}}, respectively under 1/16, 1/8, 1/4, 1/2 partition protocols. Compared to state-of-the-art methods, our method outperforms nearly all the methods in all settings.

\paragraph{Results on Cityscapes} Table \ref{t2} presents the comparison with state-of-the-art methods on the Cityscapes dataset. The UBM brings about \textcolor{red}{\textbf{+2.03, +0.58, +2.78}}, and \textcolor{red}{\textbf{+2.60}} of improvements under 1/16, 1/8, 1/4, 1/2 partition protocols compared to baseline. Also, our method outperforms all the state-of-the-art methods.  

\begin{table}[!htbp]
  \caption{Comparison with state-of-the-art methods on PASCAL VOC 2012. All the methods are based on DeepLabv3+ and ResNet-101 Backbone. All other results are referred from ST++ \citep{yang2022st++}}
  \label{t1}
  \centering
  \begin{tabular}{l c l l l l}
    \toprule
    Method   & CutMix   & 1/16(662)   & 1/8 (1323)& 1/4(2646)  & 1/2(5291) \\
    \midrule
    Supervised &  & 66.30 &70.60 & 73.10 & 77.21\\
    \midrule
    CutMix~\citep{french2019semi} &\cmark& 71.66 & 75.51 & 77.33  & 78.21  \\ 
    GCT~\citep{ke2020guided} & \xmark & 67.20  & 72.50 & 75.10 & 77.40 \\ 
    DCC~\citep{lai2021semi}  & \xmark & 72.40  & 74.60 & 76.30 & -- \\
    ST~\citep{yang2022st++} & \cmark & 72.90 & 75.70   & 76.40  &  --\\
    ST++~\citep{yang2022st++} & \cmark & \textbf{74.50} & 76.30   & 76.60 & -- \\
    \midrule
    CPS~\citep{chen2021semi}  & \xmark & 72.18 &  75.83 & 77.55  & 78.64\\
    CPS+UBM  & \xmark & $\textbf{73.59}^{\textcolor{red}{+1.41}}$ &$\textbf{77.25}^{\textcolor{red}{+1.42}}$   &$\textbf{79.30}^{\textcolor{red}{+1.75}}$ & $\textbf{80.64}^{\textcolor{red}{+2.00}}$ \\
    
    \bottomrule
  \end{tabular}
\end{table}

\begin{table}[!htbp]
  \caption{Comparison with state-of-the-art methods on Cityscapes. All methods are based on DeepLabv3+ and ResNet-101 Backbone. All other results are referred from $\text{U}^2\text{PL}$ \citep{wang2022semi}.}
  \label{t2}
  \centering
  \begin{tabular}{l c l l l l l}
    \toprule
    Method    & CutMix & 1/16(186) & 1/8(372) & 1/4(744)  & 1/2(1488)\\
    \midrule
    Supervised & & 65.74 & 72.53 & 74.43 & 77.83 \\
    \midrule
    CutMix\citep{french2019semi} &\cmark  & 67.06 & 71.83  & 76.36 & 78.25 \\ 
    GCT\citep{ke2020guided}    & \xmark & 66.75 & 72.66  & 76.11 & 78.34 \\  
    DCC~\citep{lai2021semi}  & \xmark & --  & 69.70 & 72.70 & 77.50 \\
    RCC~\citep{zhang2022region}  & \cmark & --  & 74.04 & 76.47 & -- \\
    $\text{U}^2\text{PL}$\citep{wang2022semi} & \cmark  &  70.30 &  74.37 &  76.47 & 79.05  \\
    \midrule
    CPS\citep{chen2021semi}    &  \xmark & 69.78   &  74.31 & 74.58  & 76.81  \\
    CPS+UBM & \xmark & $\textbf{71.81}^{\textcolor{red}{+2.03}}$    & $\textbf{74.89}^{\textcolor{red}{+0.58}}$  & $\textbf{77.36}^{\textcolor{red}{+2.78}}$ & $\textbf{79.41}^{\textcolor{red}{+2.60}}$  \\
    
    \bottomrule
  \end{tabular}
\end{table}

\subsection{Ablation Studies}

All the ablation studies are carried out on PASCAL VOC 2012 dataset with labeled ratio 1/4, with DeepLabv3+ and ResNet-101 backbone, the size of the vicinity is set to 15 × 15. 

\paragraph{The Effectiveness of Components in Uncertainty Booster}
To further analyze the effective portions of our methods, we separately introduce Uniform Distribution instead of RUV and remove UAS. The results are shown in Table \ref{t3} (Van. indicates Vanilla; UD indicates Uniform Distribution booster). We can see that if we add UAS for each unlabeled image, there will be an increase of \textbf{+1.49}. But if we add a Uniform Distribution booster and UAS to boost uncertainty, there will be a remarkable decrease in mIoU with \textbf{-6.39}. This proves that the distribution of pixels in different images is remarkably different. When we add both the UAS and RUV, we achieve the highest performance of \textbf{+1.93}. That is because RUV catches non-local distributions of the pixel, thus can better generate a more similar distribution to the input unlabeled image distribution.



\paragraph{Ablation Study on The Size of The Vicinity} We also ablate the vicinity size of the 2D cross-correlation operator. As table \ref{t4} shows, the results remain almost the same, which proves that if we focus on image-wise distribution for each pixel when boosting uncertainty for the model, the hyperparameters count for little influence on the model.

\begin{minipage}{\textwidth}
        \begin{minipage}[t]{0.5\textwidth}
            \centering
            \makeatletter\def\@captype{table}\makeatother\caption{Ablations on Two Strategies}
            \label{t3}
            \begin{tabular}{lcccc}
\toprule
   &Van.  &UAS & UD/UAS  & RUV/UAS     \\
    \midrule
    mIoU & 77.55 &79.04 & 72.91     & \textbf{79.48}           \\
    
    \bottomrule

\end{tabular}
        \end{minipage}
        \begin{minipage}[t]{0.5\textwidth}
        \centering
        \makeatletter\def\@captype{table}\makeatother\caption{Ablations on Different Vicinity Sizes}
        \label{t4}

        \begin{tabular}{lcccc}
\toprule
    Size      & ${3\times3}$   & ${5\times5}$ & ${9\times9}$  & ${15\times15}$ \\
    \midrule
    mIoU & 79.45  &79.30 &79.40   & \textbf{79.48} \\
    \bottomrule
\end{tabular}
        \end{minipage}
    \end{minipage}

 \begin{wrapfigure}{r}{0.5\textwidth}
 \vspace{-20pt}
 	\centering
 	\includegraphics[width=0.48\textwidth]{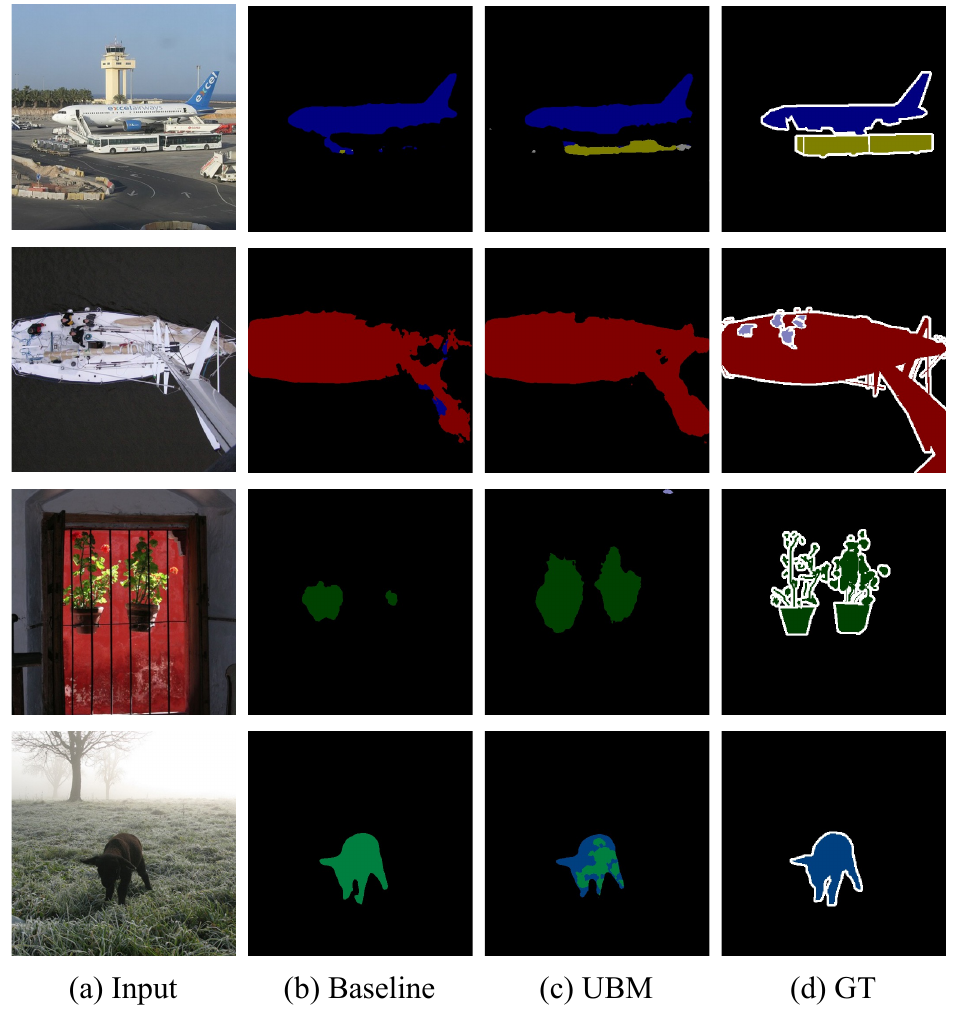}
 	\vspace{-7pt}
 	\caption{Qualitative results on PASCAL VOC 2012 val set.}
 	\label{fig:voc-vis}
 	\vspace{-43pt}
 \end{wrapfigure}

\subsection{Qualitative Results}

The qualitative results tested on 1/8 labeled data of PASCAL VOC are presented in Figure \ref{fig:voc-vis}. Our method outperforms the baseline in many scenarios. We get a more complete and accurate segmentation result rather than baseline method, and incorrect semantics can be corrected by nearby semantic information. More results are presented in the supplementary material.

\section{Conclusion and Future Outlook}

In this paper, we theoretically and experimentally propose that boosting uncertainty on unlabeled data helps with the generalization of the model in semi-supervised semantic segmentation. We demonstrate two advanced strategies to design a novel uncertainty booster. The first strategy aims to map the uncertainty-boosted output closer to the prior labeled output of the model. The second strategy proposes that the model should pay more attention to the uncertain data points, which means the more the model is unconfident, the more we boost the uncertainty of the data points. Following the theoretical strategies, we design a plug-and-play module that does not need any training. Our module makes the old baseline method outperform the current methods on PASCAL VOC 2012 and Cityscapes via different partition protocols without increasing too much training cost. Our work can trigger the research interest in the distribution gap and inspire more work on developing uncertainty methods in semi-supervised learning.

\bibliography{iclr2023_conference}
\bibliographystyle{iclr2023_conference}

\appendix
\section{Appendix}

For the vanilla model $h$, the upper bound of $ R_{D_U}^E(h) - R_{D_U}^G(h)$ turns out to be:
\begin{equation}
\label{11}
  R_{D_U}^E(h) - R_{D_U}^G(h) \leq \sqrt{\frac{\mathcal{\mathbf{KL}}[Q \| P]+\log \frac{2 \sqrt{N}}{\delta}}{2 N}}
\end{equation}

Thus, for the upper bound of Eq. \ref{11}, the K-L divergence is:

\begin{equation}
\label{12}
\begin{aligned}
\mathbf{KL}[Q \| P] &=\sum_{i=1}^{d} \mathbf{KL}(\mathcal{N}(\mathbf{W}_{lu} \bar{x}_{u}, \mathbf{I}) \| \mathcal{N}(\mathbf{W}_l \bar{x_l}, \mathbf{I})) \\
&=d\|\mathbf{W}_l \bar{x_l}-\mathbf{W}_{lu} \bar{x}_{u}\|_{2}^{2}
\end{aligned}
\end{equation}

Thus, for the vanilla model $h$, the form of Eq. \ref{11} is written as:

\begin{equation}
\label{13}
   R_{D_U}^E(h) \leq R_{D_U}^G(h) + \sqrt{\frac{d\|\mathbf{W}_l \bar{x_l}-\mathbf{W}_{lu} \bar{x}_{u}\|_{2}^{2}}{2N}} + 
   \sqrt{\frac{\log {\frac{2 \sqrt{N}}{\delta}}}{2 N}}
\end{equation}

which is shown in the main body of the paper.

\end{document}